\documentclass[conference]{IEEEtran}
\IEEEoverridecommandlockouts
\usepackage{amsmath,amssymb,amsfonts}
\usepackage{graphicx}
\usepackage{textcomp}
\usepackage{xcolor}
\usepackage{hyperref}
\usepackage{enumitem}
\usepackage{tikz}
\usepackage{lipsum}
\usepackage{caption}
\usepackage{algpseudocode}
\usepackage{multirow}
\usepackage{booktabs}
\usepackage{algorithm}
\newcommand*\circled[1]{\tikz[baseline=(char.base)]{
            \node[shape=circle,draw,inner sep=0.5pt] (char) {#1};}}
\def\BibTeX{{\rm B\kern-.05em{\sc i\kern-.025em b}\kern-.08em
    T\kern-.1667em\lower.7ex\hbox{E}\kern-.125emX}}
\usepackage{cite}
\begin{document}


\title{Cooperative Students: Navigating Unsupervised Domain Adaptation in Nighttime Object Detection}

\author{Jicheng Yuan \\
\IEEEauthorblockA{\textit{Open Distributed Systems (ODS), TU Berlin} \\
\textit{and BIFOLD Berlin}\\
Berlin, Germany \\
jicheng.yuan@tu-berlin.de}
\\
Anh Le-Tuan \\
\IEEEauthorblockA{\textit{Open Distributed Systems (ODS)} \\
\textit{TU Berlin}\\
Berlin, Germany \\
anh.letuan@tu-berlin.de}
\and
Manfred Hauswirth \\
\IEEEauthorblockA{\textit{Open Distributed Systems (ODS), TU Berlin} \\
\textit{and Fraunhofer FOKUS}\\
Berlin, Germany \\
manfred.hauswirth@tu-berlin.de}
\\
Danh Le-Phuoc \\
\IEEEauthorblockA{\textit{Open Distributed Systems (ODS), TU Berlin} \\
\textit{and Fraunhofer FOKUS}\\
Berlin, Germany \\
danh.lephuoc@tu-berlin.de}
}

\maketitle

\begin{abstract}
Unsupervised Domain Adaptation (UDA) has shown significant advancements in object detection under well-lit conditions; however, its performance degrades notably in low-visibility scenarios, especially at night, posing challenges not only for its adaptability in low signal-to-noise ratio (SNR) conditions but also for the reliability and efficiency of automated vehicles. To address this problem, we propose a \textbf{Co}operative \textbf{S}tudents (\textbf{CoS}) framework that innovatively employs global-local transformations (GLT) and a proxy-based target consistency (PTC) mechanism to capture the spatial consistency in day- and night-time scenarios effectively, and thus bridge the significant domain shift across contexts. Building upon this, we further devise an adaptive IoU-informed thresholding (AIT) module to gradually avoid overlooking potential true positives and enrich the latent information in the target domain. Comprehensive experiments show that CoS essentially enhanced UDA performance in low-visibility conditions and surpasses current state-of-the-art techniques, achieving an increase in mAP of 3.0\%, 1.9\%, and 2.5\% on BDD100K, SHIFT, and ACDC datasets, respectively. Code is available at this \href{https://github.com/jichengyuan/Cooperitive_Students}{\textit{GitHub repository}}.
\end{abstract}


\maketitle
    
\section{Introduction}
\label{sec:intro}
Object detection has made significant advancements with the advent of deep learning techniques~\cite{zou2023object}. 
While available approaches perform exceptionally well under standard lighting conditions, they often falter in nighttime or low-light scenarios. The differences in illumination, shadows, and contrasts between daytime and nighttime contexts pose significant challenges for models serving for autonomous driving, particularly in environments such as rural areas or highways where artificial lighting is sparse. Previous studies~\cite{hong2021crafting, liu2022image} primarily focused on supervised learning with synthetic data or semi-supervised domain adaptation, where labeled data from the target domain is a prerequisite. 
\begin{figure}[t]
\begin{center}
\includegraphics[width=1.0\linewidth]{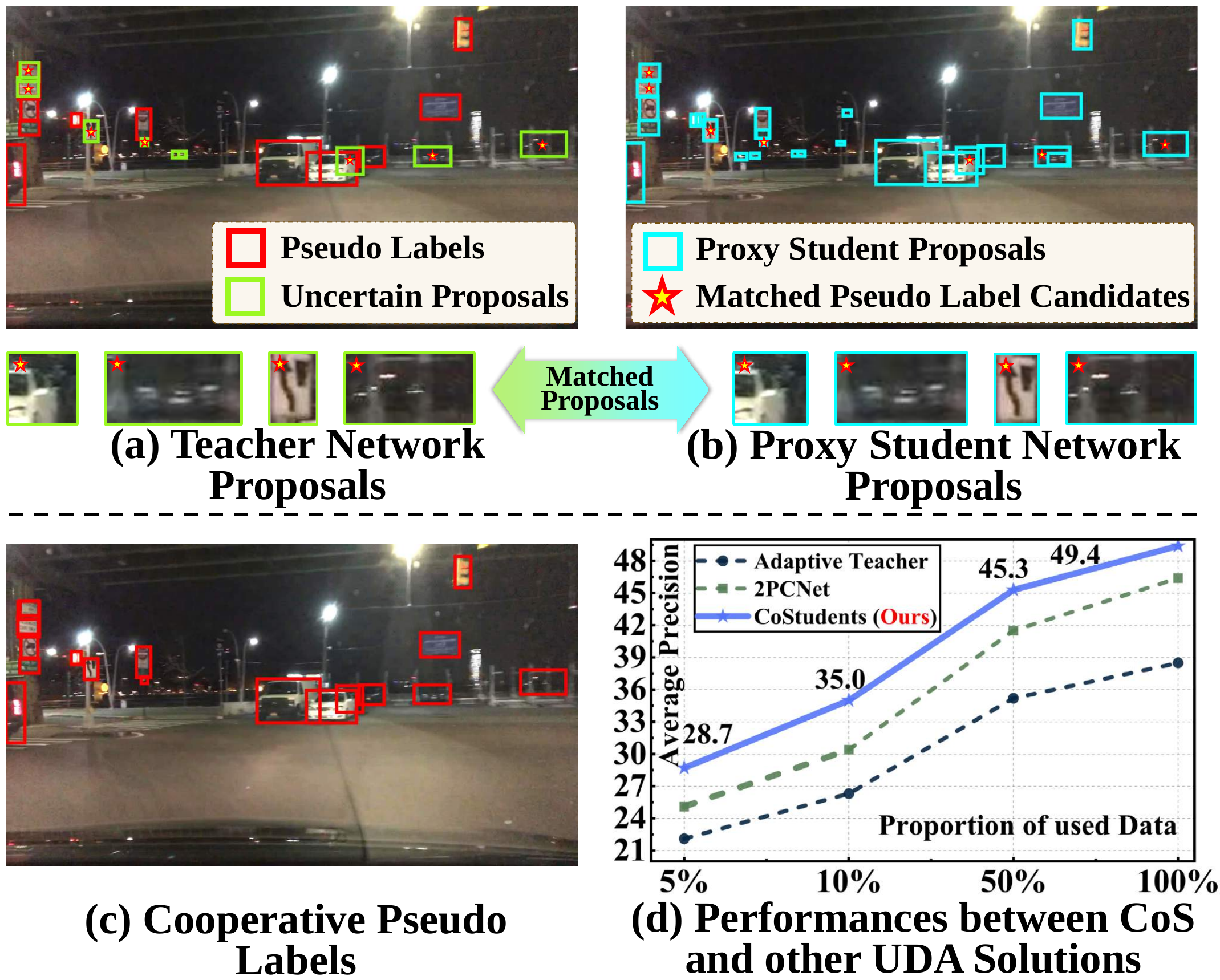}
\end{center}
\setlength{\abovecaptionskip}{1pt}
\caption{
Differences between CoS and other UDA methods. (a) Pseudo-labels (PLs) and uncertain proposals in traditional solutions. (b) and (c) PLs proposed by CoS. (d) Performance comparisons on BDD100K~\cite{yu2020bdd100k} with varying data usage. More examples are shown in the supplement. (The image was enhanced for better visibility.) 
}
\vspace{-6mm}
\label{fig:data_proportion}
\end{figure} \\
However, obtaining labeled data for every possible domain is resource-intensive and often impractical. This has led to the emergence of unsupervised domain adaptation (UDA) as a potential solution. The fundamental premise of UDA in day-to-night scenarios is to utilize models trained on daytime images and adapt them to nighttime conditions without the need for human- or model-driven annotations in the target domain. In this paradigm, the target domain benefits from latent information learned by the detector trained on the source domain. Yet, this is not without its challenges: \circled{1} A key observation in UDA for object detection is the variance in object appearances and backgrounds across domains~\cite{chen2018domain}. While some methods~\cite{mancini2018boosting, wang2018visual} focus on aligning feature distributions between the source and target domains, others~\cite{sun2022prior, ainam2021unsupervised} emphasize pseudo-labeling techniques to iteratively guide the model learning on the target domain. \circled{2} Due to the large domain gap, pseudo-labels (PLs) produced by the source domain detector for the target (nighttime) domain often lack the requisite accuracy, struggling to identify and localize objects with desired qualities. Current UDA methods~\cite{deng2021unbiased, kennerley20232pcnet} produced PLs focus predominantly on the confidence derived from the classification branch, often neglecting potential localization information or inherent uncertainty in challenging yet prone to be overlooked objects, as demonstrated in Figure~\ref{fig:data_proportion} (a). \circled{3} Relying solely on a static threshold focusing on classification confidence for PLs filtering often overlooks the fact that the model’s prediction confidence varies across categories and iterations, leading to inconsistent targets and adversely impacting the performance~\cite{wang2023consistent, chen2022dense}. Thus, the adaptive generation of high-quality PLs and the avoidance of overlooking potential true positives (TPs) are crucial for robust adaptation from daytime to nighttime contexts.
\\
In this work, we propose \textbf{Co}operative \textbf{S}tudents (\textbf{CoS}), 
a novel mutual-learning-based UDA 
framework for object detection from daytime- to nighttime- contexts, leveraging the teacher-student architecture~\cite{tarvainen2017mean}. 
To address the challenge~\circled{1}, we introduce a parameter-free global-local transformation (GLT) module to enhance daytime images with shared prior knowledge like lighting and contrast information in nighttime scenarios, thus guiding the detector to capture semantically consistent object features across domains. Against the challenge~\circled{2}, we incorporate a proxy-based target consistency (PTC) module,  leveraging mutual consistency in both classification and localization branches between teacher- and proxy-student-network, aiming to iteratively refine the learned latent information and pseudo-label qualities used to guide the student network advancing in nighttime images, as depicted in Figure~\ref{fig:data_proportion} (b) and (c). 
Additionally, building upon PTC, we develop an adaptive IoU-informed thresholding (AIT) strategy to expand the potential searching space for consistent positive samples, addressing the challenge~\circled {3}. Through these, we aim to establish a more coherent domain adaptation, resulting in enhanced performance in UDA for nighttime object detection.
Our contributions are summarized as follows:
\begin{itemize}[leftmargin=*, itemsep=2pt] 
\item {We develop a GLT module that enhances daytime images utilizing prior knowledge from nighttime scenarios, serving as a basic adaptation unit aiming at reducing the domain gap without losing the semantic relevance.}
\item {We introduce a PTC module that incorporates classification and localization information to capture overlooked consistent targets, iteratively refining the learning on nighttime images and avoiding the neglect of valuable predictions.}
\item {Based on PTC, we further adopt an AIT strategy to widen the potential searching space of true positives.}
\item Extensive experiments verify that CoS outperforms existing unsupervised methods in day-to-night adaptation under various proportions of used data, as shown in Figure~\ref{fig:data_proportion} (d). 
\end{itemize}
    
\section{Related Work}
\noindent
\textbf{Unsupervised Domain Adaptation} 
leverages labeled data from the source domain to train models, adapting them to the target domain and achieving better performance without accessing its annotations. Traditional approaches~\cite{mancini2018boosting, wang2018visual, long2017deep} aim to minimize domain discrepancies through feature alignment. Conversely, \cite{tzeng2017adversarial, li2022cross} employs a domain discriminator for mining domain-agnostic features. Others~\cite{yang2020fda, zheng2020forkgan, Awais_2021_ICCV}, either employing synthetic data or adversarial examples to reduce the domain gap, are effective in UDA. However, these solutions are resource-intensive. Different from them, we develop a parameter-free GLT module, enhancing images in the source domain by leveraging prior knowledge from the target domain.
\vspace{-1mm}
\\
\textbf{UDA in Nighttime Contexts
} 
has gained significant traction in dealing with tasks in low-visibility scenarios with limited or without annotations, whereby most research focuses on semantic segmentation~\cite{xu2021cdada, wu2021dannet} and object tracking~\cite{ye2022unsupervised}. DANNet~\cite{wu2021dannet} introduces a DA network tailored for nighttime semantic segmentation without needing labeled nighttime data. In aerial tracking, UDAT~\cite{ye2022unsupervised} employs a transformer-based bridging layer as a feature discriminator between domains, training the daytime model adversarially for nighttime adaptation. For object detection, AugGAN~\cite{lin2020gan} incorporates synthetic datasets to bridge the day-night disparity. 2PCNet~\cite{kennerley20232pcnet} introduces a two-phase consistency to mitigate domain shift. While these UDA solutions are effective, they primarily rely on classification confidence in producing PLs. In contrast, our approach uniquely emphasizes mutual consistency, combining classification and localization to mine consistent targets with semantic relevance across contexts, thus reducing the divergence between domains.
    
\section{Methodology}
\begin{figure*}[t]
\begin{center}
\includegraphics[height=2.7in]{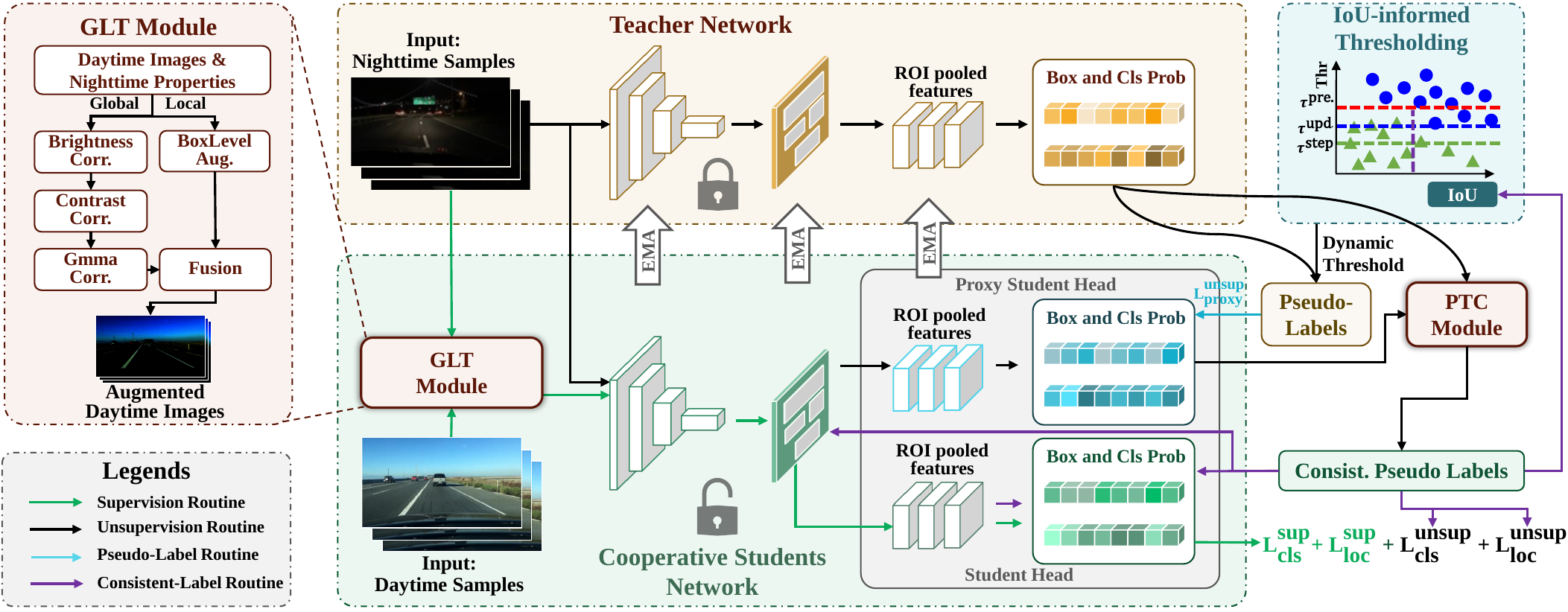}
\end{center}
\setlength{\abovecaptionskip}{1pt}
\vspace{2mm}
\caption{
Network Architecture of the proposed CoS: The key components include the Global-Local Transformation module, serving to reduce the domain gap while maintaining semantic relevance between source and target through prior knowledge from the target domain; Proxy-based target consistency module works to capture overlooked consistent targets; and adaptive IoU-informed thresholding strategy expands the searching space for potential true positives.
During domain adaptation, the corresponding source domain data will used for initial training and the target domain data will be used as inputs during the unsupervised learning stage. The refined label will be used to supervise the second student network (the bottom one). Both subnetworks have the same structure that has components: RoI and detection head with shared RPN and feature extractor. The unsupervised losses, $ \mathcal{L}^{unsup}_{cls}$ and $ \mathcal{L}^{unsup}_{loc}$  are considered between the bottom student network and the consistent PLs $\mathcal{\hat{P}}_{t}^{T_p}$ produced by teacher and proxy student networks.
}
\setlength{\belowcaptionskip}{-5mm}
\vspace{-3mm}
\label{fig:overview}
\end{figure*}
\subsection{Overview and Problem Definition}
UDA in nighttime object detection poses greater challenges than conventional UDA (under well-fit conditions) due to the sharp illumination shifts, divergent feature quality, and low SNR~\cite{lin2020gan}. Different from prior  approaches~\cite{li2022cross,kennerley20232pcnet}, CoS eases these challenges by capturing semantic relevance across domains and localization consistency of potential true positives (TPs) using a mutual-learning-based teacher-student network (TSN). Define $\mathcal{D}_s = \{\mathcal{I}_s^{N_s}, \mathcal{B}_s^{M}, \mathcal{C}_s^{M}\}$ as $N_s$ samples from the source domain, with $\mathcal{B}_s^M \in \mathbb{R}^{M \times 4}$ as bounding boxes, and $\mathcal{C}_s^M$ 
as class labels of $M$ box-level annotations. Conversely, the target domain, characterized by nighttime imagery, is denoted as $\mathcal{D}_t = \{\mathcal{I}_t^{N_t}\}$.
To ensure the quality of pseudo-labels (PLs) in the early stage UDA, images from the source domain are utilized for the initial training of the student network. The supervised loss $\mathcal{L}_{sup}$ comprises the losses from the RPN network and RoI head as: 
\begin{equation}
\min_{\Theta_s} \sum_i^{N_s} \left[\mathcal{L}_{rpn}(f_s(\mathcal{I}^i_{sa}), \mathcal{B}^i_s))+\mathcal{L}_{roi}(f_s(\mathcal{I}^i_{sa}, \mathcal{B}^i_s, \mathcal{I}^i_{sa}))\right],
\end{equation}
where $f_s\left(\cdot ; \Theta_s\right)
$ is the student detector, and $\mathcal{I}_{sa}$ denotes images enhanced by the GLT module, as detailed in Section~\ref{subsec:GLT}.
Following the supervised warm-up training, we initialize the teacher network using parameters from the student network. In the subsequent UDA phase, the teacher model produces PLs for $\mathcal{D}_t$, guiding the learning of the student network. As training advances, the quality of PLs will be progressively enhanced. However, a prevalent issue arises, as uncertain proposals depicted in green boxes in Figure~\ref{fig:data_proportion} (a), potential targets are intentionally excluded due to their classified confidences not reaching a predefined rigid threshold $\tau_{se}^{cls}$. To address this, we propose a proxy student network to ensure localization consistency for those overlooked potential TPs. Specifically, when identical category proposals emerge from both networks for a specific potential object, even if their confidences fall below the threshold $\tau_{se}^{cls}$, target consistency can be achieved based on the precision of mutual localizations, as the proposals with red $\star$ depicted in Figure~\ref{fig:data_proportion} (a), (b). Those matched proposals are then incorporated as extra candidates for PLs. 
Subsequently, the extended PLs $\mathcal{\hat{P}}_{t}^{T_p}(\mathcal{\hat{B}}_{t}^{T_p}, \mathcal{\hat{C}}_{t}^{T_p})$, collaboratively validated by the teacher and proxy-student networks, are employed to guide the learning of the student network (bottom part of Figure~\ref{fig:overview}). They serve to minimize the unsupervised loss in the target domain:
\begin{equation}
\min_{\Theta_s} \sum_i^{N_t} \left[\mathcal{L}_{cls}(f_s(\mathcal{I}_t^{i}), \mathcal{\hat{C}}_t^{{T_p}_i}))+\mathcal{L}_{reg}(f_s(\mathcal{I}_t^{i}), \mathcal{\hat{B}}_t^{{T_p}_i})\right],
\end{equation}
Additionally, we utilize the Exponential Moving Average (EMA)~\cite{tarvainen2017mean} to transfer the knowledge learned by the student detector $f_s\left(\cdot ; \Theta_t\right)
$ back to the teacher detector $f_t\left(\cdot ; \Theta_t\right)
$, as defined in Equation~\ref{formula:ema}:
\begin{equation}
\Theta_t=\alpha \cdot\Theta_{t-1}+(1-\alpha)\cdot \Theta_s,
\label{formula:ema}
\end{equation}
where $\Theta_t$ and $\Theta_s$ denote learnable parameters from the teacher and student network. Furthermore, to expand the searching space in the target domain and enhance detector performance as learning advances, we introduce an AIT module to adjust $\tau_{se}^{cls}$ dynamically, as detailed in Section~\ref{subsec:STD}. This contrasts with prior works~\cite{deng2021unbiased, li2022cross} that require a static threshold to filter valuable outputs as PLs.
\subsection{Global-Local Transformation \label{subsec:GLT}}
Given our objective of performing UDA from daytime to nighttime contexts, the disparity between these domains, particularly in aspects like luminance conditions and visibility, is substantial. UDAT~\cite{ye2022unsupervised} introduced a domain discriminator to mitigate the large domain gap between source and target. However, it increases model complexity and can lead to negative adaptation~\cite{zhang2022free}. Drawing inspiration from~\cite{tian2017global,kennerley20232pcnet}, we introduce a parameter-free method, named global-local-transformation (GLT), to transfer the prior knowledge from source to target. For global brightness transformation, we first calculate the channel-wise mean and variance, denoted as $\mu_c^n$, $\sigma_c^n$ for nighttime images, and $\mu_c^d$, $\sigma_c^d$ for the daytime image to be enhanced. Next, the luminance offset is defined as $\Delta \mu_c = \mu_c^n - \mu_c^d$ and the adaptive parameter is given by $v_c = \beta \cdot \Delta \mu_c$, where $\beta \sim \mathcal{U}(0.2, 0.8)$ simulates the nighttime luminance variability. Subsequently, we adjust the brightness channel-wise in the daytime image, as Equation~\ref{eq:adaptive_calculation} outlined:
\begin{equation}
\begin{gathered}
\mathcal{I}_{sa}^c(x, y) = \mathcal{I}_s^c(x, y)  \cdot \max(0.2, \min(v_c, 1.0)),
\label{eq:adaptive_calculation}
\end{gathered}
\end{equation}
Similarly, we conduct contrast and gamma correction to daytime images. Additionally, it is crucial to note that beyond the evident inter-domain gaps, substantial intra-domain discrepancies exist, including fluctuations in illumination and shadow presence on objects. Specifically, brightness inconsistencies can manifest across various regions of the same object captured at nighttime, influenced by factors like localized dim lighting or intense directional illumination. Hence, relying solely on global transformation can lead to regional feature discrepancies between nighttime and enhanced daytime images. To address this, we propose a box-level enhancement technique. In daytime images, we randomly set a pixel coordinate inside bounding boxes as the center point, around which a mask is positioned, either horizontally or vertically. Then, we reapply the brightness correction to this randomly mask area. The entire  GLT pipeline is illustrated in the upper left block of Figure~\ref{fig:overview}.
\begin{figure}[t]
\begin{center}
\includegraphics[width=1.0\linewidth]{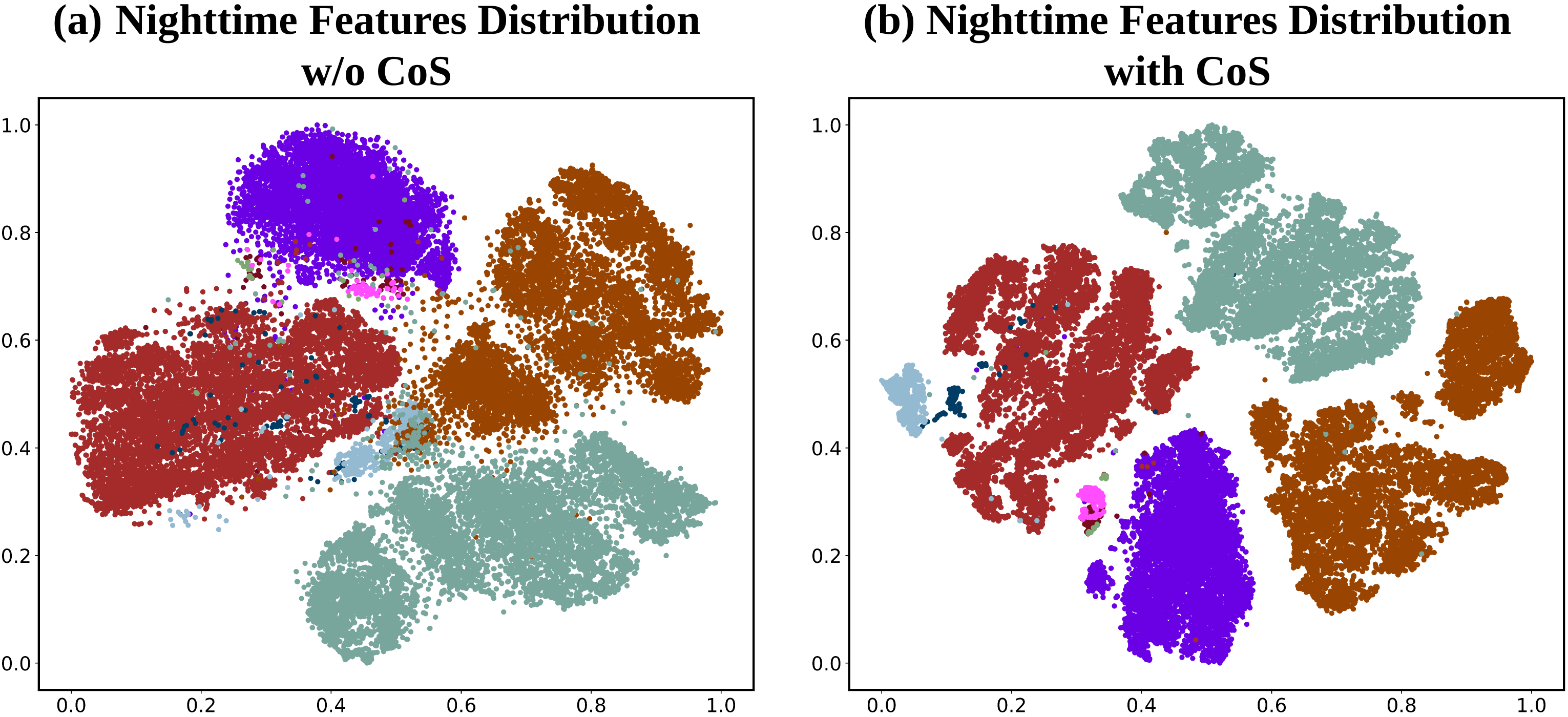}
\end{center}
\setlength{\abovecaptionskip}{1pt}
\caption{
t-SNE~\cite{van2008visualizing} visualization of nighttime object features on BDD100K~\cite{yu2020bdd100k}.}
\vspace{-5mm}
\label{fig:tsne}
\end{figure}
\subsection{Proxy-based Target Consistency \label{subsec:PTC}}
In unsupervised learning using teacher-student-network (TSN)~\cite{tarvainen2017mean}, pseudo-labels (PLs) are typically produced by the teacher based on classification confidences and a static secure threshold $\tau_{se}^{cls}$~\cite{li2022cross, deng2021unbiased}. However, as learning progresses, reliance solely on confidence scores cannot guarantee the quality of PLs and model performance in identifying potential true positives (TPs)~\cite{wang2023consistent}, as illustrated in Figure~\ref{fig:data_proportion} (a). This implies that some TPs, accurately classified and localized by the teacher, may be inadvertently overlooked due to failing to meet the predefined $\tau_{se}^{cls}$. To address this, we develop a novel proxy-based target consistency mechanism to automatically identify those consistent targets that were overlooked during learning. Specifically, a proxy student network (in the middle part of Figure~\ref{fig:overview}), bridging the traditional teacher and student network, promotes the model to distinguish potential true positives in the overlooked predictions that fall below the secure threshold $\tau_{se}^{cls}$, thus achieving target consistency among those uncertain predictions using auxiliary classification and localization information. As illustrated in Figure~\ref{fig:overview}, our method comprises three main steps: Initially, during the UDA phase on nighttime images, we feed input data $\mathcal{I}_t$ into the teacher network $f_t\left(\cdot ; \Theta_t\right)
$ to get their target outputs $\mathcal{O}_t^T$. 
Based on the $\mathcal{O}_t^T$ and secure threshold $\tau_{se}^{cls}$, the filtered initial PLs $\mathcal{\hat{P}}_t^T$ can be obtained according to the Formula~\ref{eq:pseudo_label_init}.
\begin{equation}
\mathcal{\hat{P}}_t^T(\mathcal{\hat{B}}_t^T, \mathcal{\hat{C}}_t^T) \leftarrow
\mathcal{O}_t^{T}(\mathcal{B}_t^{T}, \mathcal{C}_t^{T}) \quad where \quad \mathcal{C}_t^{T_i} \geq \tau_{se}^{cls},
\label{eq:pseudo_label_init}
\end{equation}
Similarly, the same nighttime input $\mathcal{I}_t$ without annotations is fed into the proxy student network to obtain outputs $\mathcal{O}_t^{P}$. The initial PLs $\mathcal{\hat{P}}_t^T$ produced by the teacher network will guide the learning of the proxy student model, as $\mathcal{L}_t^p= \frac{1}{N_t}\sum_i^{N_t} \left[\mathcal{L}_{cls}(f_p(\mathcal{I}_t^{i}), \mathcal{\hat{C}}_t^{{T}_i}))+\mathcal{L}_{reg}(f_p(\mathcal{I}_t^{i}), \mathcal{\hat{B}}_t^{{T}_i})\right]$. Next, outputs $\mathcal{O}_t^{T}$ from the teacher and $\mathcal{O}_t^{P}$ from the proxy student will be aggregated to produce second-stage PLs based on mutual classification and localization information, achieving target consistency. Specifically, the consistent targets will be mined between the uncertain (overlooked) predictions (from the teacher) and outputs $\mathcal{O}_t^{P}$ (from the proxy student). The final mined consistent targets will be considered as pseudo-label candidates $\mathcal{\hat{P}}_{t}^{P}$ if two bounding boxes (one from teacher and another from proxy student) share identical category assignments and their localization consistency exceeds $\tau_{se}^{loc}$ (set at 0.8 in this work), as described in 
Formula~\ref{eq:pseudo_label_cs}. 
\begin{equation}
\begin{gathered}
\mathcal{\hat{P}}_{t}^{P}(\mathcal{\hat{B}}_{t}^{P}, \mathcal{\hat{C}}_{t}^{P}) = \mathcal{{O}}_{t}^T(\mathcal{{B}}_{t}^T, \mathcal{{C}}_{t}^T) \otimes \mathcal{{O}}_{t}^P(\mathcal{{B}}_{t}^P, \mathcal{{C}}_{t}^P),
\\
\quad where \quad \text{argmax}(\mathcal{C}_t^{T_i}) = \text{argmax}({C}_{t}^{P_i}), \quad \mathcal{C}_t^{T_i} < \tau_{se}^{cls} 
\\
\quad and \quad \mathcal{F}_{loc}(\mathcal{{B}}_{t}^{T_i}, \mathcal{{B}}_{t}^{P_i}) \geq \tau_{se}^{loc}
\label{eq:pseudo_label_cs}
\end{gathered}
\end{equation}
Then, $\mathcal{\hat{P}}_t^T$ and $\mathcal{\hat{P}}_{t}^{P}$ are merged to form the final consistent pseudo-labels $\mathcal{\hat{P}}_{t}^{T_p}$, as defined in Formula~\ref{eq:pseudo_label_final},
\begin{equation}
\begin{gathered}
\mathcal{\hat{P}}_{t}^{T_p}(\mathcal{\hat{B}}_{t}^r, \mathcal{\hat{C}}_{t}^r) = \mathcal{\hat{P}}_t^T(\mathcal{\hat{B}}_t^T, \mathcal{\hat{C}}_t^T) \oplus \mathcal{\hat{P}}_{t}^P(\mathcal{\hat{B}}_{t}^P, \mathcal{\hat{C}}_{t}^P),
\label{eq:pseudo_label_final}
\end{gathered}
\end{equation}
guiding the learning of the student network, as depicted at the bottom of Figure~\ref{fig:overview}. 
To guarantee the quality of the consistent PLs $\mathcal{\hat{P}}_{t}^{T_p}$ produced by teacher and proxy student networks, we use a warm-up phase: the PTC module will be introduced after 40\% of iterations in the burn-up stage. During the warm-up phase, we follow the training setup in~\cite{deng2021unbiased}, solely the initial PLs $\mathcal{\hat{P}}_{t}^{T}$ generated by the teacher model are used to supervise the student network.
\begin{table*}[htb]
\centering
\scalebox{1.0}{
\begin{tabular}{@{}lcccccccccc|c@{}}
\toprule
\multicolumn{2}{c}{Methods} & Bicycle & Bus & Car & Rider & Truck & Motorcycle & Pedestrian & Traffic Light & Traffic Sign & AP \\ 
\midrule
\multicolumn{2}{l}{\textit{Oracle:}} & & & & & & & & & & \\
 & Night-time & 36.2 & 42.5 & 73.2 & 27.4 & 47.5 & 27.2 & 51.0 & 45.6 & 66.6 & 41.7 \\
 & Day-\&Night-time & 42.7 & 51.6 & \textbf{75.5} & \underline{34.3} & \textbf{56.1} & 36.3 & \underline{56.0} & \textbf{51.8} & \underline{66.8} & \underline{47.1} \\
\midrule
\multicolumn{2}{l}{\textit{I2I based:}} & & & & & & & & & & \\
 & CycleGAN~\cite{zhu2017unpaired} & 43.1 & 52.0 & 69.9 & 34.9 & 50.1 & 34.8 & 52.3 & 33.0 & 62.6 & 43.2 \\
 & UNIT~\cite{liu2017unsupervised} & 39.1 & 51.7 & 70.4 & 34.9 & 50.1 & 33.2 & 52.2 & 37.5 & 62.7 & 43.3 \\
 & ForkGAN~\cite{zheng2020forkgan} & 39.4 & 50.3 & 69.2 & 29.6 & 48.7 & 32.5 & 49.9 & 44.6 & 61.8 & 42.6 \\
 & FourierDA~\cite{yang2020fda} & 38.2 & 49.0 & 69.4 & 26.5 & 44.2 & 30.2 & 46.2 & 43.2 & 61.5 & 41.8 \\
\midrule
\multicolumn{2}{l}{\textit{Enhance based:}} & & & & & & & & & & \\
 & DAF~\cite{chen2018domain} & 30.3 & 37.4 & 56.5 & 25.3 & 35.8 & 21.5 & 35.8 & 25.8 & 44.5 & 31.3 \\
 & TDD~\cite{he2022cross} & 25.9 & 35.6 & 68.4 & 20.7 & 33.3 & 16.5 & 43.1 & 43.1 & 59.5 & 34.6 \\
 & SADAF~\cite{chen2021scale} & 33.0 & 35.1 & 67.7 & 24.3 & 38.6 & 33.0 & 49.7 & \underline{51.5} & 61.4 & 38.4 \\
 & Adaptive Teacher~\cite{li2022cross} & 42.7 & 52.1 & 60.8 & 30.4 & 48.9 & 34.5 & 42.3 & 29.1 & 43.9 & 38.5 \\
 & Unbiased Mean Teacher~\cite{deng2021unbiased} & 40.2 & 46.3 & 46.8 & 26.1 & 44.0 & 28.2 & 46.5 & 31.6 & 52.7 & 36.2 \\
 & Two Phase Consistent Network~\cite{kennerley20232pcnet} & \underline{44.5} & \underline{55.2} & 73.1 & 30.8 & 53.8 & \underline{37.5} & 54.4 & 49.4 & 65.2 & 46.4 \\
\midrule
&\textbf{Cooperative Students (Ours)} & \textbf{50.2} & \textbf{55.6} & \underline{73.9} & \textbf{41.3} & \underline{54.7} & \textbf{43.5} & \textbf{57.1} & 50.0 & \textbf{67.8} & \textbf{49.4} \\
\bottomrule
\end{tabular}
}
\caption{Performance comparison (in AP) of Cooperative Students with other approaches for day-to-night UDA. Results are shown for the supervised model (oracle), image-to-image translation (I2I-based), and image enhancement (enhancement-based) solutions on the BDD100K dataset. Bold represents the best performance, and underline indicates the second-best performance within each category.}
\label{tab:sota}
\end{table*}

\subsection{Adaptive IoU-Informed Thresholding}\label{subsec:STD}
Prior studies~\cite{chen2018domain,chen2021scale,he2022cross} employed a static hyperparameter $\tau_{se}^{cls}$ to filter potential samples as PLs. However, this overlooks the variation in the model performance across iterations, leading to missing potential true positives and significantly impacting the overall learning progress.
For instance, a high $\tau_{se}^{cls}$ can result in too many false negatives, while a low value may increase false positives, compromising model performance. To ease this, building upon the PTC module, we propose an Adaptive IoU-informed thresholding (AIT) strategy to adjust the $\tau_{se}^{cls}$ dynamically and capture potential TPs as learning advances. It adjusts the threshold by analyzing the consistent PLs derived from PTC and focuses on those samples that, even if they do not meet stringent classification thresholds, are significantly relevant in localization consistency, as illustrated in the upper right of Figure~\ref{fig:overview}. Specifically, with AIT, the mean $\mu_t^P$ and variance $\sigma_t^P$ are first calculated using the classification confidences above their $median(\cdot)$ scores in matching boxes between the teacher and proxy student network. Then, an auxiliary threshold update step is introduced: $\tau_{stp}^{cls}=\mu_t^P-2\times\sigma_t^P$. Next, The deviation for threshold adjustment is derived as: $\tau_{dev}^{cls}=\gamma(\tau_{stp}^{cls}-\tau_{se}^{cls})$. Here, $\gamma$ is an adjustable parameter set by default at 0.05. The updated threshold, defined as $\tau_{upd}^{cls} = \tau_{se}^{cls} + \tau_{dev}^{cls}$, is employed in subsequently iterations. Hence, as training progresses, the threshold is incrementally lowered to capture more valuable true positives, benefiting from the enhanced model performance.
    
\section{Experiments}
\begin{table}[] 
\centering
\scalebox{1.0}{
\label{tab:ablation}
\large
\resizebox{0.48\textwidth}{!}{
\begin{tabular}{c|c|c|c|cccc}
\toprule[1.5pt]
\multirow{2}{*}{\begin{tabular}[c]{@{}c@{}}Target Consistent\\ Module\end{tabular}} & 
\multirow{2}{*}{\begin{tabular}[c]{@{}c@{}}Global\\ Trans.\end{tabular}} & \multirow{2}{*}{\begin{tabular}[c]{@{}c@{}}Local\\ Trans.\end{tabular}} & \multirow{2}{*}{\begin{tabular}[c]{@{}c@{}}Threshold\\ Decay\end{tabular}} & \multirow{2}{*}{$AP$} & 
\multirow{2}{*}{$AP^L$} & 
\multirow{2}{*}{$AP^M$} & 
\multirow{2}{*}{$AP^S$} \\ 
& & & & & & & \\ \hline

& & & & 35.1 & 32.3 & 18.6 & 5.7 \\
\hline
$\textbf{\textit{}}\checkmark$ & & & & 41.3 & 38.4 & 23.3 & 7.9 \\
$\textbf{\textit{}}\checkmark$ & & $\textbf{\textit{}}\checkmark$& & 44.0 & 39.6 & 19.3 & 8.4 \\

$\textbf{\textit{}}\checkmark$ & $\textbf{\textit{}}\checkmark$ & & & 47.9 & 42.2 & 26.8 & 9.2 \\

$\textbf{\textit{}}\checkmark$ & $\textbf{\textit{}}\checkmark$ & $\textbf{\textit{}}\checkmark$ & & 48.4 & 44.8 & 27.2 & 9.6  \\
\hline
$\textbf{\textit{}}\checkmark$ & $\textbf{\textit{}}\checkmark$ & $\textbf{\textit{}}\checkmark$ & $\textbf{\textit{}}\checkmark$ & 49.4 & 45.9 & 27.9 & 10.2 \\ 
\bottomrule[1.5pt]
\end{tabular}}}
\caption{Ablation study on BDD100K
~\cite{yu2020bdd100k}.}
\vspace{-4mm}
\label{tab:ablation}
\end{table}
\subsection{Datasets and Evaluation Metrics}
We conducted extensive experiments on CoS to assess its effectiveness in real-world and synthetic scenarios across BDD100K~\cite{yu2020bdd100k}, SHIFT~\cite{sun2022shift}, and ACDC~\cite{sakaridis2021acdc} (More details about datasets are in the supplement).
Additionally, ablation studies were performed to evaluate the contribution of each module under various configurations of $\tau_{se}^{cls}$, $\tau_{se}^{loc}$ in PTC and $\gamma$ in AIT. All experiments reported mean average precision (mAP) with an evaluation threshold of 0.5.
\subsection{Implementation Details}
We adopted FRCNN~\cite{ren2015faster} with pre-trained ResNet-50 as the primary object detector. For the PTC module, initial secure confidence thresholds for classification $\tau_{se}^{cls}$, and localization $\tau_{se}^{loc}$, were set at 0.8 in all experiments, unless otherwise specified. Additionally, in AIT, the decay setting $\gamma$ for $\tau_{se}^{cls}$ is 5e-2, applied post 60\% completion of the burn-up stage.
During the learning phase, the student network was initiated with a learning rate of 4e-2. The first 50k iterations form the burn-in stage, focusing on the knowledge transfer from source to target. In the next 50k iterations, the burn-up stage dedicated to the UDA phase, we set the EMA for updating the teacher model at 0.9996. The PLT module is introduced post 40\% of the burn-up stage iterations to enhance PLs robustness and capture more potential TPs. All the experiments were conducted on two Nvidia A40 GPUs, with each GPU handling a batch size of nine. The implementation was carried out using the PyTorch framework.
\begin{figure}[t]
\begin{center}
\includegraphics[width=1.0\linewidth]{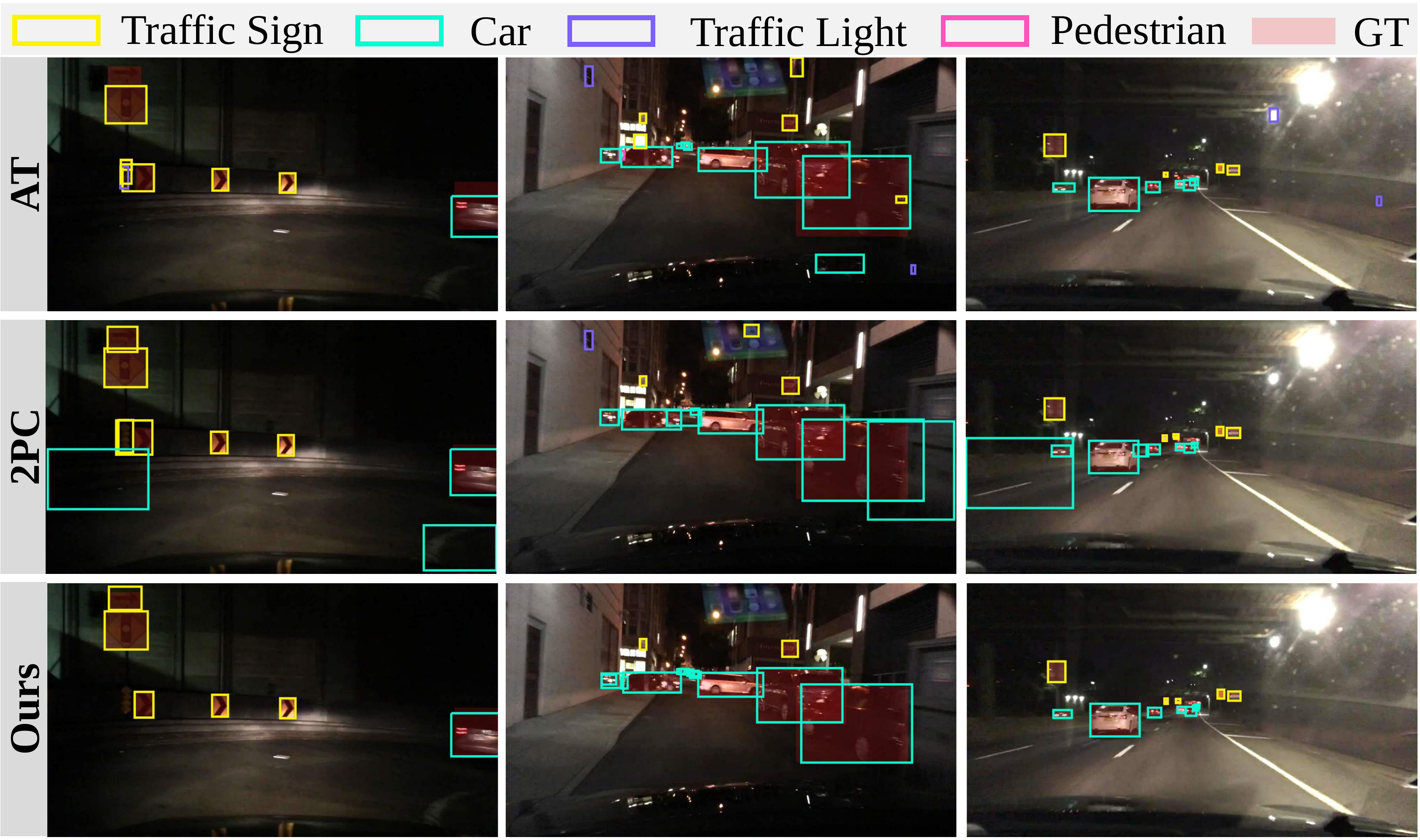}
\end{center}
\setlength{\abovecaptionskip}{1pt}
\caption{
Qualitative comparison of CoS with other methods. Images were enhanced for improved visibility. Additional examples are available in the supplement.}
\vspace{-2.5mm}
\label{fig:dets}
\end{figure}
\subsection{Comparison to the state of the art}
As shown in Table~\ref{tab:sota}, we firstly evaluated CoS against leading state-of-the-art image enhancement~\cite{yang2020fda, li2022cross, kennerley20232pcnet} and image2image (I2I) translation approaches~\cite{zheng2020forkgan, yang2020fda, zhu2017unpaired, pang2021image} on the BDD100K~\cite{yu2020bdd100k}. Owing to the sparse presence of the \textit{train} category in BDD100K, we excluded it from all configurations. Impressively, the proposed CoS demonstrates superior performance, exhibiting approximately a 3.0\% increase in AP compared to its rank-1 competitor~\cite{kennerley20232pcnet} and nearly a 2.3\% improvement over the fully supervised method, especially in small and challenging classes, such as \textit{bicycle}, \textit{rider} and \textit{motorcycle}. Furthermore, Figure~\ref{fig:dets} show that CoS consistently outperforms in diverse real-world illumination conditions (with and without artificial lighting in low visibility scenarios), whereas other approaches, AT~\cite{li2022cross} and 2PC~\cite{kennerley20232pcnet}, either inaccurately detect dark-areas or artificial lighting as false-positives or miss objects (false-negatives) under low illumination (additional results are in the supplement). This underscores the performance improvement of leveraging consistent predictions identified by CoS, no matter how the illumination conditions and the size of objects vary. Additionally, we extracted features from RoI and used t-SNE to visualize them. Figure~\ref{fig:tsne} shows that without CoS, while the model partially aligns the marginal distribution in some classes, it fails to clearly discriminate between categories. However, CoS effectively aligns distributions and discriminates between categories. This confirms the capacity of CoS in transferring knowledge across domains, effectively executing UDA in object detection for nighttime context.
Given the limited availability of real-world data, we evaluated CoS on the synthetic dataset SHIFT~\cite{sun2022shift}. Table~\ref{tab:SHIFT} (left) indicates that CoS surpasses the oracle model with a 1.5\% increase in AP. Furthermore, we evaluate CoS in more challenging situations on ACDC~\cite{sakaridis2021acdc}, executing UDA solely from daytime images under adverse weather conditions to nighttime contexts. As shown in Table~\ref{tab:SHIFT} (right), CoS yields the best results, with improvements of 3.5\% and 1.9\% over the 2PC~\cite{kennerley20232pcnet} and fully supervised model, respectively. This confirms the effectiveness of CoS in both clear and adverse weather conditions, as well as on synthetic data. This stems from target consistency and collaboration between proxy-student and student networks, essential for identifying true positives and extracting valuable latent information.
\begin{table}[htbp] 
\centering
\scalebox{1.0}{
\large
\resizebox{0.48\textwidth}{!}{
\begin{tabular}{c|ccc|ccc}
\toprule[1.5pt]
Methods & 
$AP_{ST}^{.50:.95}$ & 
$AP_{ST}^{.75}$ & 
$AP_{ST}^{.50}$ & 
$AP_{AC}^{.50:.95}$ & 
$AP_{AC}^{.75}$ & 
$AP_{AC}^{.50}$   \\ 
\hline
Oracle &32.0 &34.8 &49.5 &18.1 &19.2 &37.1\\
\hline
FourierDA~\cite{zhu2017unpaired} &30.3 &32.7 &47.1 &17.0 &15.5 &34.6  \\
SADAF~\cite{chen2021scale} &30.4 &34.6 &46.4 & 16.3 & 14.0 & 34.4  \\
AT~\cite{li2022cross} &29.2 &33.0 &44.8 & 16.3 & 15.6 & 30.6 \\
2PC~\cite{kennerley20232pcnet} &31.9 &34.9 &49.1 & 18.7 & 17.3 & 36.5   \\
\hline
\textbf{Ours} &33.8 &37.3 &51.0 & 21.3 & 20.4 & 39.0 \\ 
\bottomrule[1.5pt]
\end{tabular}}}
\caption{Quantitative results on SHIFT~\cite{sun2022shift} and ACDC ~\cite{sakaridis2021acdc} datasets.}
\vspace{-3mm}
\label{tab:SHIFT}
\end{table}

\subsection{Ablation Study}
To assess the contribution of each module in CoS, we conduct ablation experiments on BDD100K~\cite{yu2020bdd100k} under various setups. As shown in Table~\ref{tab:ablation}, the initial baseline without proposed modules achieves 35.1\% AP. Then, with PTC, model performance improves by 6.2\%. Yet, relying solely on consistent targets does not well bridge the large domain gap between daytime and nighttime contexts. After incorporating the GLT module, CoS achieved 48.4\% performance in AP. This indicates that, with the integration of GLT, more valuable latent information extracted from daytime images plays a crucial role in bridging the domain gap. Considering the enhanced model performance across iterations, after introducing the AIT module, CoS reached the best performance on BDD100K with 49.4\% AP. This suggests that benefiting from the expanded searching space, CoS is capable of capturing potential TPs that were overlooked by the teacher network and thereby promoting learning progress. With the integration of GLT, PTC, and AIT modulus, CoS effectively narrows the large domain gap, achieving consistent semantic relevance between the daytime and nighttime features.
\begin{figure}[htbp]
\begin{center}
\includegraphics[width=1.0\linewidth]{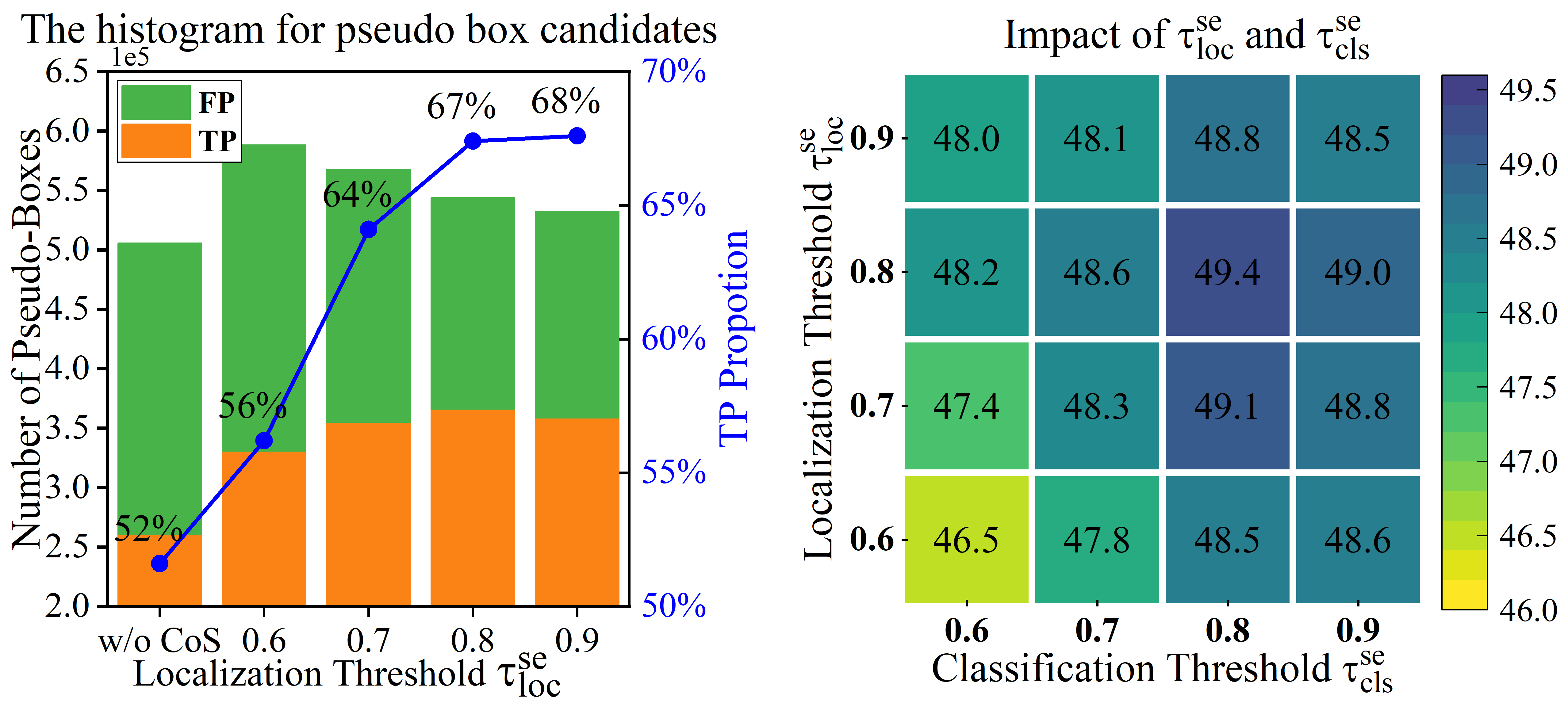}
\end{center}
\setlength{\abovecaptionskip}{1pt}
\caption{
Sensitivity analysis on $\tau^{cls}_{se}$ and $\tau^{loc}_{se}$.}
\vspace{-1mm}
\label{fig:ablation_data}
\end{figure}

To further investigate the joint impact of the secure thresholds in classification $\tau^{cls}_{se}$ and localization $\tau^{loc}_{se}$, we conducted additional experiments shown in Figure~\ref{fig:ablation_data}.
The results demonstrate that the PTC module significantly enhances the proportion of potential True Positives (TPs) as $\tau^{loc}_{se}$ increases. However, setting lower thresholds leads to a rapid increase in false positives. Meanwhile, the number of TPs in matched proposals peaks at $\tau^{loc}_{se}=0.8$. Additionally, setting both $\tau^{cls}_{se}$ and $\tau^{loc}_{se}$ at 0.8 yields the optimal performance of CoS on the BDD100K~\cite{yu2020bdd100k} dataset.
    
\section{conclusion and future work}
In this study, we introduce CoS, a novel framework designed for UDA in nighttime object detection. CoS incorporates the GLT module to align semantic features between domains, addressing illumination and feature variations. The PTC module, building upon TSN, reliably identifies potential TPs consistently, enhancing generalizability. Complementing this, the AIT module, based on PTC, expands the searching space for potential positive proposals and retains more valuable PLs, ensuring smoother domain adaptation. CoS demonstrates marked improvement in nighttime object detection and outperforms existing methods on benchmark datasets. Future work will aim to bolster CoS robustness in complex and composite conditions across weather, seasons, and cities and integrate with scalable world models to form universal domain detectors, thereby minimizing risks in rare scenarios.

\section{Acknowledgements}
This work is supported by the Deutsche Forschungsgemeinschaft, German Research Foundation under grant number 453130567 (COSMO), by the Horizon Europe Research and Innovation Actions under grant number 101092908 (SmartEdge), by the Federal Ministry for Education and Research, Germany under grant number 01IS18037A (BIFOLD).

    \bibliographystyle{bibStyle}
    \bibliography{bib}


    
\end{document}